%% file: acl_latex.tex
\documentclass[11pt]{article}
\usepackage{float}

\usepackage[final]{acl}

\usepackage{times}
\usepackage{pifont}
\usepackage{booktabs}
\usepackage{caption}
\usepackage{makecell}
\usepackage{latexsym}
\usepackage{amsmath}
\usepackage{inconsolata}
\usepackage{enumitem}

\usepackage{multirow}
\newcommand{\uls}{\begin{itemize}[leftmargin=*]}
\newcommand{\ule}{\end{itemize}}
\newcommand{\ols}{\begin{enumerate}[leftmargin=*]}
\newcommand{\ole}{\end{enumerate}}
\newcommand{\li}{\item}
\usepackage[T1]{fontenc}
\usepackage[utf8]{inputenc}

\usepackage{microtype}
\usepackage{hyperref}
\usepackage{soul}

\usepackage{inconsolata}

\usepackage{graphicx}

\title{From Scratch to Fine-Tuned: A Comparative Study of Transformer Training Strategies for Legal Machine Translation}

\author{
 \textbf{Amit Barman},
 \textbf{Atanu Mandal},
 \textbf{Sudip Kumar Naskar}
\\
 Jadavpur University, Kolkata, INDIA, 
\\
 \small{
   \textbf{Correspondence:} \href{mailto:amitbarman811@gmail.com}{amitbarman811@gmail.com}
 }
}

\begin{document}
\maketitle
\begin{abstract}
% The system created by Team JUNLP at Jadavpur University for the Shared Task on Legal Machine Translation (L-MT) of JUST-NLP 2025, which is located in collaboration with IJCNLP-AACL 2025\footnote{\url{https://www.codabench.org/competitions/10351/}}, is presented in this article. The endeavor focuses on translating legal materials, which pose considerable difficulties because of their formal linguistic expressions, complex structure, and domain-specific vocabulary. Our submission investigates: (i) using the provided domain-specific Indian legal data to fine-tune the pre-trained Helsinki-NLP/opus-mt-en-hi model; and (ii) using the same data to train a transformer-based model from scratch. With a BLEU score of 46.03 and a fourth place on the official scoreboard, the optimized OPUS model showed exceptional performance. Domain adaptation by focused fine-tuning successfully improves translation accuracy and fluency in the legal context, according to the evaluation of a number of measures, including METEOR, CHRF++, BERTScore, and COMET.
In multilingual nations like India, access to legal information is often hindered by language barriers, as much of the legal and judicial documentation remains in English. Legal Machine Translation (L-MT) offers a scalable solution to this challenge by enabling accurate and accessible translations of legal documents. This paper presents our work for the JUST-NLP 2025 Legal MT shared task, focusing on English–Hindi translation using Transformer-based approaches. We experiment with 2 complementary strategies, fine-tuning a pre-trained OPUS-MT model for domain-specific adaptation and training a Transformer model from scratch using the provided legal corpus. Performance is evaluated using standard MT metrics, including SacreBLEU, chrF++, TER, ROUGE, BERTScore, METEOR, and COMET. Our fine-tuned OPUS-MT model achieves a SacreBLEU score of 46.03, significantly outperforming both baseline and from-scratch models. The results highlight the effectiveness of domain adaptation in enhancing translation quality and demonstrate the potential of L-MT systems to improve access to justice and legal transparency in multilingual contexts.
\end{abstract}

\section{Introduction}
\input{sections/introduction}

\section{Related Works}

\input{sections/related_works.tex}

\section{Dataset Description}
% \input{sections/problem_statement}
\input{sections/dataset}

% \input{sections/experimental}

\section{Experiments}
\input{sections/experiments}

\section{Results and Analysis}

\input{sections/results}

\section{Conclusion and Future Work}
\input{sections/conclusion}

\section*{Limitations}
% Although the proposed approach shows very encouraging results in the legal translation setting, there are several limitations restricts the scope of the current work. First, our model was trained exclusively on the dataset provided by the organizers of the shared task, and no extra legal or general-domain corpus was used. Thus, the model has limited exposure to a variety of linguistic constructions and context variations for legal texts. Moreover, during the training phase, the complete validation data (source–target pairs) were not provided, which impedes proper overfitting or underfitting analysis and reduces the possibilities for model calibration through hyperparameter tuning. To make the computation possible, the input and output sequence lengths were also restricted to 128 tokens. While in most cases this was sufficient for sentence-level translations, longer legal clauses or compound sentences might fall victim to truncation, which can affect the coherence and completeness of the translation. Finally, due to time constraints and resource limitations, our study used mainly automated measures like SacreBLEU and chrF++ without qualitative or human judgment-based evaluations. Future studies should test longer input sequences, make use of larger and more diverse legal corpora, and include human judgment for assessing adequacy and fluency of the translations in real-life legal settings.
While the proposed approach demonstrates strong empirical performance, several limitations constrain the generalizability and scope of the current study:
\uls
    \li \textbf{Restricted training corpus:} The model was trained exclusively on the dataset released by the shared task organizers, without augmentation from external legal or general-domain corpora. Consequently, the system's exposure to broader linguistic variability and complex domain phenomena remains limited.
    \li \textbf{Lack of comprehensive validation data:} Complete source-target validation pairs were unavailable during training, which hindered reliable monitoring of model behavior (e.g., overfitting or underfitting) and constrained opportunities for principled hyperparameter optimization.
    \li \textbf{Sequence length constraints:} Input-output sequences were truncated to a maximum of 128 tokens due to computational limitations. Although suitable for most sentence-level examples, this restriction may adversely affect the processing of lengthy statutory clauses, compound sentences, or cross-referential structures.
    \li \textbf{Absence of human evaluation:} The assessment relies primarily on automated metrics (SacreBLEU, chrF++, BERTScore, COMET), and does not incorporate expert human judgement, limiting deeper qualitative insights into adequacy, legal constancy, and pragmatic interpretability.
    \li \textbf{Resource constraints:} Due to time and computational constraints, broader experimental exploration, such as parameter-efficient tuning, multilingual transfer, or alternative architectures-was not undertaken.
\ule

% Future work will focus on supporting longer sequences, leveraging diverse large-scale legal corpora, incorporating expert evaluation, and exploring scalable adaptation strategies for high-constancy legal translation.

\section*{Acknowledgments}
\input{sections/ack}

% Bibliography entries for the entire Anthology, followed by custom entries
%\bibliography{anthology,custom}
% Custom bibliography entries only
\bibliography{custom}

% \appendix

% \section{Example Appendix}
% \label{sec:appendix}

% This is an appendix.

\end{document}

%% file: sections/introduction.tex
Since India's independence in 1947, language has remained one of the defining features and challenges of its democracy. The Constitution recognizes 22 scheduled languages, but much of the country's legal, administrative, and judicial work continues to be conducted in English. This linguistic imbalance often leaves citizens dependent on translations to understand laws, judgments, or government notifications that affect their rights. There have been documented instances where individuals have misunderstood court proceedings or official orders simply because they were not available in their native language, an obstacle that runs counter to the ideal of ``equal access to justice''.

In a multilingual democracy, ensuring that legal information is accessible to all citizens is not only a linguistic challenge but also a civic necessity. 
% \input{sections/figure}
% \textcolor{blue}{Figure \ref{fig: Illustration} illustrates a real situation where the legal term ``suo motu'' is misunderstood due to literal translation rather than transliteration. The phrase means \textit{``on its own motion''} and should be preserved as-is in translation, yet its misinterpretation can cause significant confusion.}
Legal texts are particularly complex, they demand precision, consistency, and adherence to jurisdiction-specific terminology. Even small translation errors can lead to misinterpretations, contractual disputes, or procedural delays. As legal materials increasingly move to online platforms, the need for accurate, scalable translation tools has become even more urgent.

Advances in Neural Machine Translation (NMT) have transformed the field of translation, enabling systems to model intricate linguistic relationships and long-range dependencies through attention mechanisms \cite{Vaswani_2017}. The rise of Large Language Models (LLMs) trained on vast multilingual data has further improved translation fluency and generalization. Yet, these models often struggle in highly specialized domains like law, where vocabulary, syntax, and semantics diverge significantly from general text. Domain-specific adaptation remains essential for achieving accurate and trustworthy translations.

This paper focuses on developing Legal Machine Translation (L-MT) systems that bridge the linguistic divide in the Indian legal context. As part of the JUST-NLP 2025 Legal MT shared task\footnote{\url{https://exploration-lab.github.io/JUST-NLP/}}, we investigate how Transformer-based models can be adapted for English–Hindi legal translation. We explore two strategies, training a Transformer model from scratch and fine-tuning the OPUS-MT model, to assess how domain-focused training influences translation quality.

Through this work, we aim to advance the development of reliable and inclusive Legal MT systems that make legal information accessible across languages, supporting transparency, participation, and justice in multilingual societies.

The key contributions of this paper are summarized as follows:
\uls
    % \vspace{-0.8em}
    \li We trained and evaluated a Transformer model from scratch on legal-domain data.
    % \vspace{-0.8em}
    \li We fine-tuned the Helsinki Opus MT for legal-domain adaptation.
    % \vspace{-0.8em}
    \li We analyzed translation robustness and domain adaptability across evaluation datasets.
\ule

%% file: sections/related_works.tex
Machine Translation (MT) has long been one of the most prominent applications of Natural Language Processing (NLP). Early MT systems were primarily built upon sequence-to-sequence architectures using encoder–decoder frameworks. However, due to their sequential nature and reliance on recurrent neural networks, these models often struggled to capture long-range contextual dependencies effectively.

The introduction of the self-attention mechanism revolutionized MT by enabling models to capture global dependencies among tokens more efficiently. Transformer-based architectures have since become the foundation of modern MT systems, demonstrating exceptional generalization across languages and domains through large-scale multilingual pretraining. This paradigm shift has significantly improved translation fluency, adequacy, and semantic consistency.

Recent advancements in LLMs have further enhanced multilingual translation capabilities through zero-shot and few-shot learning. These pretrained multilingual models can produce reasonable translations even without explicit task-specific fine-tuning. However, their performance tends to degrade substantially for low-resource language pairs, where limited data hampers generalization. To address this, research has increasingly focused on fine-tuning and transfer learning strategies that enable domain and language adaptation. Techniques such as multilingual continued pretraining, cross-lingual embeddings, and parameter-efficient fine-tuning (e.g., adapters like LoRA \cite{Hu-2021} and QLoRA \cite{Dettmers-2023}) have proven effective in improving translation quality for low-resource scenarios. These methods balance computational efficiency with adaptability, allowing pretrained multilingual models to specialize in specific linguistic domains such as legal, medical, or conversational text.

In the Indian context, legal translation has emerged as a crucial area of research due to the nation's linguistic diversity and the absence of a single national language. The growing need to make legal documents accessible across India's many official languages highlights the importance of domain-specific MT systems. However, Indian languages often lack large, high-quality parallel corpora, posing challenges for training robust legal MT models \cite{joshi-2024}.

Over the past decade, several multilingual parallel corpora have been developed for Indian languages. Notable examples include Samanantar \cite{ramesh-2022}, corpus for 11 Indian languages and the corpus by \citet{siripragada-2020}, which covers 10 Indian languages. Broader evaluation was also enabled by the FLORES-200 benchmark \cite{nllbteam-2022}. Other valuable resources include IndoWordNet \cite{kunchukuttan-2020}, PMIndia \cite{haddow-2020}, and datasets such as IITB English-Hindi \cite{kunchukuttan-2018}, BUET English-Bangla \cite{hasan-2020}, English-Tamil \cite{Ramasamy-2012}, English-Odia \cite{parida-2020}, and the Mizo-English corpus \cite{haulai-2023}. However, these datasets generally pertain to general-domain translation and are not tailored to the legal domain.

In contrast, the legal domain has seen relatively limited multilingual MT resources. International initiatives such as the Europarl corpus \cite{koehn-2005}, EUR-Lex \cite{baisa-2016}, and the UN Parallel Corpus \cite{ziemski-2016}, the Bilingwis Swiss Law Text collection \cite{hofler-sugisaki-2014} have provided valuable multilingual datasets for legal proceedings in European languages. However, these resources are largely tailored to European legal systems, linguistic structures, and translation conventions, which differ substantially from the Indian legal and linguistic context. Consequently, such corpora cannot be directly leveraged for Indian-language MT tasks, where distinct terminologies, legal frameworks, and multilingual diversity necessitate domain-specific datasets and adaptation strategies.

Within India, only a handful of initiatives have attempted to build legal-domain corpora. The Hindi–Telugu legal dataset from LTRC \cite{mujadia-2022} and the Anuvaad corpus\footnote{\url{https://github.com/project-anuvaad/anuvaad-parallel-corpus}} represents early efforts, however, they lack expert validation. The recently introduced MILPaC corpus \cite{Mahapatra-2025} marks a significant advancement, offering a well-curated, expert-validated, and multilingual benchmark for legal MT. Additionally, the WMT25 Legal Domain Test Suite \cite{singh-2025} provides a robust evaluation framework for assessing MT capabilities in English–Hindi legal translation. Together, these initiatives represent an emerging but still underdeveloped ecosystem for legal-domain MT in Indian languages.

%% file: sections/dataset.tex
% \subsection{Dataset Description}
% The task organizers supplied the dataset for this study, which comprises of parallel sentence pairs between English and Hindi for machine translation in the legal sector. Only the training source-translation pairs were initially made available by the organizers; the validation and test reference translations were withheld, and only the source sentences for those splits were made available for the assessment phase. Only after the final ranking results were revealed were the appropriate Hindi reference translations for validation and testing.
Table \ref{tab_data_stat} summarizes the dataset used in this study. Provided by the task organizers, it consists of English–Hindi parallel sentence pairs from the legal domain. Only the training pairs were initially released, while validation and test references were withheld. Participants generated translations for these sets during the evaluation and final phases, with the reference translations revealed after the leaderboard announcement.

\begin{table}[ht!]
\centering
\caption{Dataset statistics for Legal Machine Translation (L-MT) Shared Task}
% \scriptsize
\small
\begin{tabular}{lccc}
\textbf{Language Pair}    & \textbf{Train} & \textbf{Validation} & \textbf{Test} \\\hline\hline
\textbf{English-Hindi} & 50,000         & 5,000               & 5,000         \\\hline\hline     
\end{tabular}

\label{tab_data_stat}
\end{table}

% \begin{table*}[h!] %\small
% \centering
% \begin{tabular}{c|cc|cc|cc}
% \multirow{2}{*}{\textbf{Dataset}} & \multicolumn{2}{c|}{\textbf{Data Sample}}      & \multicolumn{2}{c|}{\textbf{Max Length}}     & \multicolumn{2}{c}{\textbf{Mean Length}}     \\ \cline{2-7} 
%                          & \multicolumn{1}{c|}{\textbf{English}} & \textbf{Hindi}  & \multicolumn{1}{c|}{\textbf{English}} & \textbf{Hindi} & \multicolumn{1}{c|}{\textbf{English}} & \textbf{Hindi} \\ \hline
% \textbf{Train}                    & \multicolumn{1}{c|}{50,000}  & 50,000 & \multicolumn{1}{c|}{429}     & 386   & \multicolumn{1}{c|}{155}     & 149   \\ \hline
% \textbf{Validation}               & \multicolumn{1}{c|}{5,000}   & 5,000  & \multicolumn{1}{c|}{383}     & 372   & \multicolumn{1}{c|}{153}     & 148   \\ \hline
% \textbf{Test}                     & \multicolumn{1}{c|}{5,000}   & 5,000  & \multicolumn{1}{c|}{360}     & 308   & \multicolumn{1}{c|}{151}     & 131   \\ \hline
% \textbf{Total}                    & \multicolumn{1}{c|}{60,000}  & 60,000 & \multicolumn{1}{c|}{-}       & -     & \multicolumn{1}{c|}{-}       & -     \\ 
% \end{tabular}
% \caption{Descriptive statistics for the Training, Validation, and Test dataset}
% \label{tab_data_stat}
% \end{table*}

The dataset contains 60,000 English–Hindi parallel sentences from the legal domain, divided into 50,000 for training and 5,000 each for validation and testing. The wide variation in sentence length suggests diverse syntactic structures typical of legal text. While the dataset is well-balanced and cleanly split, the absence of metadata on text type or source (e.g., statutes, judgments, or general documents) limits fine-grained domain analysis.

% The table \ref{tab_data_stat} contains the complete dataset descriptions.

% The training set includes 50,000 English–Hindi parallel sentences, having maximum morpheme sequence length of 429, an average source length of 155, and maximum morpheme sequence lengths of 386, an average translation length of 149 tokens, respectively. 

% It demonstrated broad linguistic coverage with 50,000 distinct Hindi sentences and 49,722 unique English sentences.
% There were 5,000 English-Hindi sentence pairings in the validation set, with average source and translation lengths of 153 and 148 tokens, respectively, and maximum lengths of 383 and 372. There were 5,000 distinct translation sentences and 4,987 unique source sentences in this split, suggesting significant language diversity and little duplication.
% The test set, which also included 5,000 English-Hindi pairs, showed maximum lengths of 360 and 308 tokens, respectively, with an average source length of 151 tokens and an average translation length of 131 characters. There were 4,996 distinct Hindi sentences and 4,981 distinct English sentences.
% The splits were created and made public by the organizers, and the whole dataset consisted of 60,000 samples. 

%% file: sections/experiments.tex
\input{tables/bench}

\input{tables/compare}

Transformer-based architectures have become the foundation of modern NMT due to their ability to model complex contextual relationships through self-attention. They outperform traditional sequence-to-sequence models, particularly in tasks requiring structural precision and contextual awareness, which are vital in legal translation.
However, large Transformer models are computationally expensive. For this study, we employed two complementary training strategies suitable for constrained resources:
\uls
    % \vspace{-0.8em}
    \li Fine-tuning a pre-trained OPUS-MT model to adapt general translation knowledge to the legal domain, and
    % \vspace{-0.8em}
    \li Training a Transformer from scratch to evaluate its capability to learn domain-specific patterns directly from legal text.
\ule
Our experimental setup is available in the following Link\footnote{\url{https://github.com/atanumandal0491/Legal-Translation}}.

\subsection{Opus Fine-Tune}

We fine-tuned the Helsinki Opus-MT model\footnote{\href{https://huggingface.co/Helsinki-NLP/opus-mt-en-hi}{Helsinki Opus-MT}} using the provided training corpus. Since validation references were initially withheld, evaluation was based on interim submissions. The model was optimized using the AdamW optimizer with a learning rate of $2 \times 10^{-5}$, weight decay 0.01, and batch size 32. Both input and target sequences were limited to 128 tokens to ensure computational efficiency without excessive truncation. As previous research \cite{cho-2014} suggests, excessively long inputs degrade model performance due to weakened attention over long dependencies, hence, this cap provides an effective trade-off between fidelity and efficiency.

% A detailed summary of the fine-tuning results is presented in Table \ref{tab_result}.

\subsection{Transformer Training}

To evaluate the impact of learning solely from domain-specific data, we trained a compact Transformer model from scratch. The configuration included 4 encoder-decoder layers, 8 attention heads, model dimension of 128, dropout of 0.1, and token length of 256, and a vocabulary size derived from a SentencePiece tokenizer of 32,000. The model was trained with a batch size of 32, using the Adam optimizer.
Despite limited data, this model demonstrated strong convergence, underscoring the ability of smaller Transformers to effectively learn domain-specific translation patterns when carefully optimized.

\subsection{Evaluation Setup}
Model outputs were assessed using multiple metrics capturing lexical, syntactic, and semantic correspondence: SacreBLEU~\cite{bleu2002,post-2018}, chrF++~\cite{chrf-2015}, TER~\cite{snover-2006}, ROUGE~\cite{lin-2004}, BERTScore~\cite{Zhang2019}, METEOR~\cite{Lavie-2007,banerjee-2005}, and COMET~\cite{rei-2020}.
These complementary measures ensure robust evaluation across the dimensions of precision, recall, fluency, and semantic alignment.

%% file: tables/bench.tex
\begin{table*}[!htbp]
\caption{Final leaderboard results for the JUST-NLP 2025 Shared Task on Legal Machine Translation (English-Hindi). The best scores for each metric are highlighted. Our system (JUNLP) achieved Rank 4 with competitive performance across lexical and semantic metrics.}
\centering
\label{tab:rank}
\resizebox{\textwidth}{!}{
\begin{tabular}{lccccccccc}
\toprule
\textbf{Rank} & \textbf{Team Name}      & \textbf{Country}        & \textbf{BLEU $\uparrow$}                   & \textbf{chrF++ $\uparrow$}        & \textbf{TER $\downarrow$}           & \makecell{\textbf{BERTScore}\\\textbf{(F1) $\uparrow$}}    & \textbf{METEOR $\uparrow$ }           & \textbf{COMET $\uparrow$}         & \textbf{AutoRank $\uparrow$} \\
\midrule
1              & Team-SVNIT     & India          & \textbf{51.61}          & 73.29          & \textbf{37.09}          & 92.61          & \textbf{75.80}           & \textbf{76.36}          & \textbf{61.62} \\
2              & FourCorners    & Thailand       & 50.19          & \textbf{73.67}          & 42.32          & \textbf{92.70}           & 69.54          & 75.74          & 60.31 \\
3              & goodmen        & India          & 48.56          & 73.07          & 41.63          & 92.38          & 67.15          & 75.16          & 59.39 \\
4              & JUNLP          & India          & 46.03$^6$          & 70.59$^4$          & 42.08$^3$          & 91.19$^4$          & 71.84$^3$          & 73.72$^4$          & 58.90 \\
5              & JUST-MEI       & India          & 46.67          & 70.03          & 44.63          & 90.86          & 72.86          & 72.12          & 58.79  \\
6              & Lawgorithms    & India          & 46.27          & 68.32          & 43.06          & 91.03          & 71.80           & 72.14          & 58.26 \\
7              & Tokenizers     & India          & 34.08          & 56.75          & 55.25          & 87.39          & 61.78          & 65.20           & 50.87\\
\bottomrule
\end{tabular}
}
\end{table*}

%% file: tables/compare.tex
\begin{table*}[!ht]
\centering
\caption{Comparison of translation performance across different models on the English-Hindi legal dataset.}
\label{tab:results}
\resizebox{\textwidth}{!}{
\begin{tabular}{lccccccccccc}
\toprule
\textbf{Model} & \textbf{Fine-Tuned} & \textbf{BLEU $\uparrow$} & \textbf{chrF++ $\uparrow$} & \textbf{TER $\downarrow$} & \textbf{ROUGE-1 $\uparrow$} & \textbf{ROUGE-2 $\uparrow$} & \textbf{ROUGE-L $\uparrow$} & \makecell{\textbf{BERTScore}\\\textbf{(F1) $\uparrow$}} & \textbf{METEOR $\uparrow$} & \textbf{COMET $\uparrow$} \\
\midrule
\makecell[l]{OPUS-MT\\(fine-tuned)} & \ding{51} & \textbf{46.03} & \textbf{70.59} & \textbf{42.08} & \textbf{72.42} & \textbf{52.63} & \textbf{69.05} & \textbf{91.19} & \textbf{71.84} & \textbf{73.72} \\ 
\makecell[l]{OPUS-MT\\(baseline)} & \ding{55} & 9.39 & 27.66 & 83.40 & 36.30 & 13.38 & 32.93 & 76.91 & 30.25 & 50.80 \\
\makecell[l]{Transformer\\(trained from scratch)} & \ding{55} & 37.77 & 60.88 & 59.72 & 35.98 & 13.62 & 35.69 & 88.37 & 65.58 & 64.29 \\
\makecell[l]{NLLB\\(3.3B distilled)} & \ding{55} & 23.72 & 47.50 & 63.29 & 49.00 & 26.31 & 45.78 & 85.14 & 45.32 & 67.25 \\
IndicTrans2 & \ding{55} & 10.87 & 42.36 & 81.25 & 37.89 & 11.07 & 37.10 & 81.21 & 41.78 & 60.38 \\
\bottomrule
\end{tabular}
}
\end{table*}

%% file: sections/results.tex
Table~\ref{tab:rank} presents the final leaderboard results from the JUST-NLP 2025 Shared Task on Legal Machine Translation, comparing the performance of participating systems across a range of lexical, semantic, and edit-based evaluation metrics. Our system, JUNLP, achieved an overall Rank 4, with a SacreBLEU score of 46.03, chrF++ of 70.59, and TER of 42.08, demonstrating strong translation accuracy, requiring relatively low post-editing effort. The model also performed competitively in semantic evaluation, achieving a BERTScore (F1) of 91.19, METEOR of 71.84, and COMET of 73.72, indicating high alignment with human reference translations. While the best-performing team attained marginally higher results across several metrics, our system performed in the mid-range compared to the other participating systems (cf. Table~\ref{tab:results}), underscoring the effectiveness of domain-focused fine-tuning for legal translation.

Table~\ref{tab:results} summarizes our experimental outcomes, comparing the fine-tuned OPUS-MT models with baseline multilingual models. The baseline OPUS-MT (without fine-tuning) performed poorly, with a SacreBLEU of 9.39 and chrF++ of 27.66, revealing significant deviation from reference translations. BERTScore F1 of 76.91 and a COMET score of 50.8 further indicate weak semantic alignment and limited adaptability of the baseline OPUS-MT to the legal domain.

The fine-tuned OPUS-MT markedly improved translation quality, achieving a SacreBLEU of 46.03, chrF++ of 70.59, and TER of 42.08, demonstrating high lexical accuracy and fluency. The BERTScore (91.19) and COMET (73.72) show strong semantic alignment with human references, while METEOR (71.84) and ROUGE scores confirm consistent n-gram and paraphrase correspondence. These results suggest that fine-tuning effectively transfers linguistic and contextual knowledge from general corpora to specialized legal data without overfitting. This performance reinforces the viability of fine-tuning for domain-specific translation and motivates further exploration of scalable approaches such as parameter-efficient tuning and extension to additional Indian languages.

The Transformer model trained from scratch performed competitively, achieving a SacreBLEU of 37.77 and COMET of 64.29. Despite lacking pre-trained initialization, it captured domain patterns effectively, although the fine-tuned OPUS-MT maintained an edge in fluency and semantic coherence. Multilingual baselines, such as NLLB and IndicTrans2, performed moderately, underscoring that general-purpose models struggle with domain-specific precision.

Overall, the fine-tuned OPUS-MT model produced fluent, accurate, and contextually faithful translations, confirming its effectiveness for English–Hindi legal MT in real-world settings.

%% file: sections/conclusion.tex
This work explored domain adaptation strategies for Legal Machine Translation (L-MT) in the English-Hindi context, highlighting how fine-tuning enhances translation quality for specialized text. Among all systems tested, the fine-tuned OPUS-MT model achieved the highest performance, demonstrating superior lexical accuracy and semantic consistency. Training a Transformer model from scratch also yielded promising results, showing that domain-specific supervision alone can produce competitive results under constrained resources.

Future work will extend these experiments to other Indian languages and evaluate parameter-efficient fine-tuning techniques such as LoRA and QLoRA to scale Legal MT further. Ultimately, such systems can play a transformative role in democratizing access to legal knowledge, ensuring that linguistic diversity does not become a barrier to justice.

%% file: sections/ack.tex
This research was funded by the `VIDYAAPATI: Bidirectional Machine Translation Involving Bengali, Konkani, Maithili, Marathi, and Hindi' under the Project titled `National Language Translation Mission (NLTM): BHASHINI'.